\documentclass[letterpaper]{article} 
\usepackage{aaai25}  
\usepackage{times}  
\usepackage{helvet}  
\usepackage{courier}  
\usepackage[hyphens]{url}  
\usepackage{graphicx} 
\usepackage{natbib}  
\usepackage{caption} 
\usepackage{algorithm}
\usepackage{algorithmicx}
\usepackage{algpseudocode}
\usepackage{amsmath}
\usepackage{booktabs} 
\usepackage{multirow}
\usepackage{subfigure}
\usepackage{color}
\usepackage{amsfonts}

\usepackage{newfloat}
\usepackage{listings}
\DeclareCaptionStyle{ruled}{labelfont=normalfont,labelsep=colon,strut=off} 
\floatstyle{ruled}
\newfloat{listing}{tb}{lst}{}
\floatname{listing}{Listing}
\title{C2F-TP: A Coarse-to-Fine Denoising Framework for Uncertainty-Aware Trajectory Prediction}
\author {
    Zichen Wang\textsuperscript{\rm 1},
    Hao Miao\textsuperscript{\rm 2},
    Senzhang Wang\textsuperscript{\rm 1}\thanks{Corresponding author},
    Renzhi Wang\textsuperscript{\rm 1},
    Jianxin Wang\textsuperscript{\rm 1},
    Jian Zhang\textsuperscript{\rm 1}
}
\affiliations {
    Central South University\textsuperscript{\rm 1}\\
    Aalborg University\textsuperscript{\rm 2}\\
    zcwang0422@csu.edu.cn, haom@cs.aau.dk, szwang@csu.edu.cn, \\
    rzwang516@gmail.com, jxwang@mail.csu.edu.cn, jianzhang@csu.edu.cn
}

\begin{document}
\maketitle
\begin{abstract}
Accurately predicting the trajectory of vehicles is critically important for ensuring safety and reliability in autonomous driving. Although considerable research efforts have been made recently, the inherent trajectory uncertainty caused by various factors including the dynamic driving intends and the diverse driving scenarios still poses significant challenges to accurate trajectory prediction. To address this issue, we propose C2F-TP, a coarse-to-fine denoising framework for uncertainty-aware vehicle trajectory prediction. C2F-TP features an innovative two-stage coarse-to-fine prediction process. Specifically, in the spatial-temporal interaction stage, we propose a spatial-temporal interaction module to capture the inter-vehicle interactions and learn a multimodal trajectory distribution, from which a certain number of noisy trajectories are sampled. Next, in the trajectory refinement stage, we design a conditional denoising model to reduce the uncertainty of the sampled trajectories through a step-wise denoising operation. Extensive experiments are conducted on two real datasets NGSIM and highD that are widely adopted in trajectory prediction. The result demonstrates the effectiveness of our proposal.
\end{abstract}

\begin{links}
    \link{Code}{https://github.com/wangzc0422/C2F-TP}
\end{links}

\section{Introduction}
Vehicle trajectory prediction aims to predict the future trajectory of the target agent based on the historical trajectories of itself and its surrounding neighbours. Accurately predicting the future trajectory of vehicles is crucial for many autonomous driving applications, including optimal driving path planning, making accurate driving decisions in dynamic environments, and enhancing driving safety~\cite{survey, lun2024resisting}.

\begin{figure}[t]
    \centering
    \includegraphics[width=1.0\linewidth]{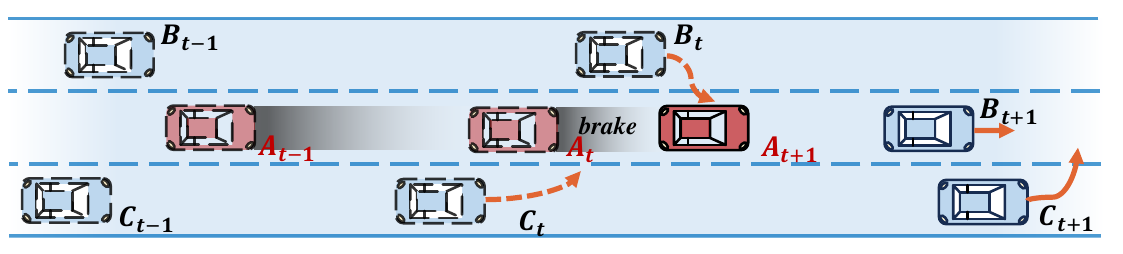}
    \caption{Illustration of the dynamics and temporal correlation of inter-vehicle interactions.}
    \label{fig1}
\end{figure}

Traditionally, statistical models are employed to predict future trajectories based on the historical trajectories of individual agents~\cite{svm, Gaussian}. Nonetheless, they do not consider the interactions between the target agent and the surrounding agents, which deteriorates the prediction performance. To address this issue, various deep learning based models~\cite{Sociallstm,cslstm, trajectron++, pip, tnt} are proposed to model the spatial interactions among vehicles. However, existing deep learning based approaches fall short in modeling the uncertainty of the trajectories~\cite{stddn,wsip}, which is common in real-world driving scenarios~\cite{cdstin}. Generally, uncertainty in autonomous driving can be broadly categorized into uncertainty in driving behavior and in driving scenarios~\cite{ten}. How to design an uncertainty-aware trajectory prediction model is still not fully explored and remains as an open research problem.

However, achieving the uncertainty-aware trajectory prediction is non-trivial due to the following challenges. 
First, the inherent uncertainty in trajectories caused by various uncertainty factors in dynamic driving scenarios, make the accurate future trajectory prediction difficult~\cite{multi}. Existing approaches mostly directly predict the coordinate points of future trajectories~\cite{Sociallstm,cf} without considering such uncertainty factors. Although some recent methods are proposed for multimodal trajectory prediction~\cite{cslstm,stddn}, they still cannot effectively model the complex uncertainty.
Second, it is challenging to simultaneously model the dynamics and temporal correlations of vehicle interactions. Existing approaches either only learn a fixed interaction weight between two vehicles~\cite{Sociallstm,cslstm} or simply consider the dynamics of interactions by modeling them at each timestamp~\cite{wsip}. For example, as shown in Figure~\ref{fig1}, at time $t$, vehicle $A$ is driving at a constant speed, and vehicles $B$ and $C$ intend to accelerate to merge into the middle lane. To avoid a collision, vehicle $A$ may slow down to keep a safe distance from vehicles $B$ and $C$, and vehicle $C$ may delay the lane-changing considering vehicle $B$'s lane-changing behavior at time $t+1$. Third, it is challenging to model the multimodal driving intentions. Most existing methods~\cite{cslstm,wsip, miao2024task} perform trajectory prediction by considering multiple intentions simultaneously. However, different motion intentions have diverse impacts on the future trajectories. Simply combining them together may lead to a suboptimal performance.

To address these challenges, we propose a coarse-to-fine denoising framework named C2F-TP for uncertainty-aware trajectory prediction. C2F-TP features a novel two-stage prediction process including a spatial-temporal interaction module and a refinement module. Specifically, in the spatial-temporal interaction module, C2F-TP aims to learn a multimodal future trajectory distribution based on historical trajectories. Next, a certain number of noisy trajectories are sampled from the learned distribution and fed into the refinement module. The refinement module adopts a conditional denoising model to generate stable and reliable future trajectories through stepwise denoising, greatly reducing the uncertainty of the input trajectories.
In particular, in the spatial-temporal interaction module, motivated by~\cite{wsip} we propose an interaction pooling to simultaneously capture the dynamic and temporal correlations of interactions across vehicles. 
To adaptively learn the impact of different motion modes on trajectory prediction, we propose a re-weighted multi-modal trajectory predictor to embed each motion mode separately. The learned embeddings are then fused adaptively by considering their diverse importance to future trajectory prediction. Moreover, we learn a multimodal future trajectory distribution based on the fused embeddings.
Finally, we sample $k$ coarse-grained trajectories from the trajectory distribution and feed them into the refinement stage. The refinement module uses historical trajectories as a condition to refine these $k$ coarse-grained trajectories, reducing their uncertainty through stepwise denoising and improving the accuracy of trajectory prediction.

The primary contributions of this paper are as follows.
\begin{itemize}
\item We for the first time propose a novel coarse-to-fine trajectory prediction model with a two-stage generation process, which considers the complex interactions across vehicles in trajectory prediction.
\item To capture the complex spatial and temporal interactions, we design a spatial-temporal interaction module with a re-weighted multi-modal trajectory predictor to model the sophisticated trajectory distributions. 
\item To contend with the inherent stochastic nature of trajectories, we design a trajectory refinement stage based on the denoising diffusion model.
\end{itemize}

\section{Related Work}
\noindent\textbf{Trajectory Prediction.} 
Trajectory prediction attracts increasing interest due to the increasing availability of trajectory data~\cite{highd, cslstm}. 
Recently, significant advances have been achieved in developing deep learning based methods~\cite{survey} for trajectory prediction. First, RNNs~\cite{analysis, wang2020deep} have been applied to predict the future movement paths of agents with historical trajectory data. However, in complex traffic environments, surrounding vehicles may significantly affect the prediction result due to the perception of interactions between agents. To address this issue, many recent studies~\cite{Sociallstm, socialstgcnn, stddn,wsip, LightTR} try to model the spatial interactions across agents by means of various emerging pooling techniques. Despite their advances, existing methods suffer from effective data uncertainty capturing.

\noindent\textbf{Denoising diffusion probabilistic models.} 
Denoising diffusion probabilistic models have achieved advanced performance on various applications~\cite{dpm, ddpm}, such as image generation~\cite{image1} and audio synthesis~\cite{audio}. The diffusion model is first proposed by DPM~\cite{dpm}, which seeks to imitate the diffusion process in non-equilibrium statistical physics and reconstruct the data distribution using the denoising model. Then, a line of studies are proposed to improve the efficiency of the diffusion model by developing fast sampling techniques, such as DDPM~\cite{ddpm} and DDIM~\cite{ddim}. Recently, the diffusion model has been applied to trajectory data, which obtain good performance in trajectory-related applications, such as trajectory synthesis~\cite{zhu2024difftraj} and trajectory prediction~\cite{led}. However, these methods fall short in effectively modeling the interactions across agents.

\section{Problem Definition and Preliminaries}

\textbf{Problem Definition.} We consider trajectory prediction as a sequence generation problem which generates the future trajectory of an agent based on the historical trajectories of itself and surrounding agents. For a target agent, given its historical trajectories $\boldsymbol{\mathrm{X_{tar}}} = \left\{{X}_{tar}^{t-T_{h}}, \ldots, {X}_{tar}^{t-2}, {X}_{tar}^{t-1}\right\}$ and trajectories of surrounding $N$ agents $\boldsymbol{\mathbb{X}_{\mathcal{N}}}= \left\{\boldsymbol{\mathrm{{X}_{1}}}, \boldsymbol{\mathrm{{X}_{2}}}, \ldots, \boldsymbol{\mathrm{{X}_{i}}}, \ldots, \boldsymbol{\mathrm{{X}_{N}}}\right\}$ over $T_h$ time steps , we aim to predict its future trajectory ${\boldsymbol{\mathrm{Y_{tar}}}}=\left\{{Y}_{tar}^{t+1}, {Y}_{tar}^{t+2}, \ldots, {Y}_{tar}^{t+T_{f}}\right\}$ over $T_f$ future time steps, where ${X}_{v}^{i}$ is the two-dimensional coordinate $({x}_{v}^{i}, {y}_{v}^{i})$ of the agent $v$ at the $i$-$th$ time step.

\textbf{Preliminaries.}
\begin{figure}
    \centering
    \includegraphics[width=0.95\linewidth]{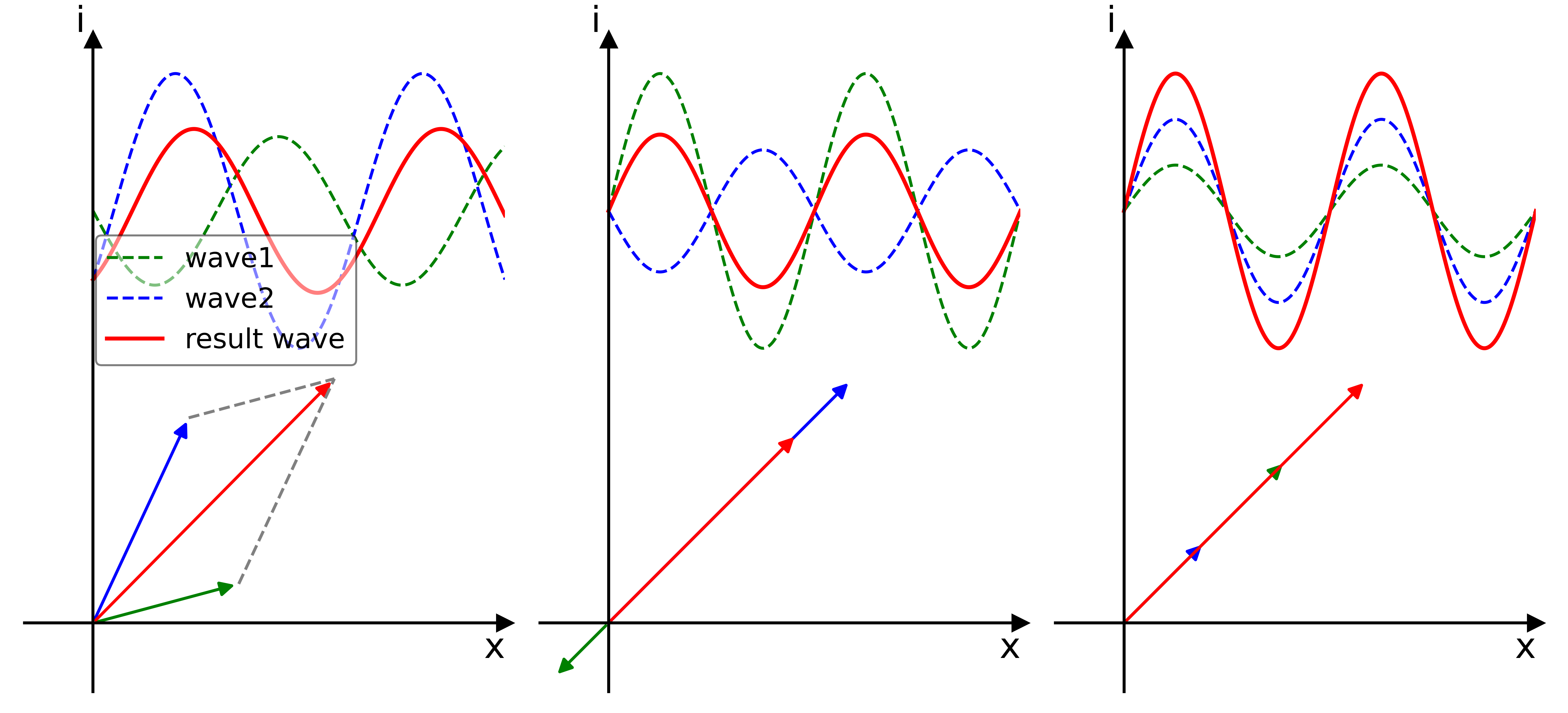}
    \caption{Interaction between two waves with different phases. Below is a superposition of two waves in the complex-valued domain, and the figure above shows how their projections along the real axis vary with phase.}
    \label{fig:wave}
\end{figure}
In quantum mechanics, an entity (\emph{e.g.}, electron, photon) is usually represented by a wave function (\emph{e.g.}, de Broglie wave) containing both amplitude and phase~\cite{wave}. Inspired by Wave-MLP~\cite{wavemlp}, an agent is represented as a wave $\tilde{z}_{j}$ with both amplitude and phase information, i.e.,
\begin{equation}
  \tilde{\boldsymbol{z}}_{j}=\left|\boldsymbol{z}_{j}\right| \odot e^{i \boldsymbol{\theta}_{j}}, j=1,2, \cdots, n,  
  \label{eqa 1}
\end{equation}
where $i$ is the imaginary unit, $\odot$ is element-wise multiplication. The amplitude $\left|\boldsymbol{z}_{j}\right|$ represents the content of each agent. $\boldsymbol{\theta}_{j}$ indicates the phase, which is the current location of an agent within a wave period. With both amplitude and phase, each agent $\tilde{\boldsymbol{z}}_{j}$ is represented in the complex-value domain. We view agent interaction as a superposition of waves, supposing $\tilde{z}_{r}=\tilde{z}_{i}+\tilde{z}_{j}$ is the aggregated results of agent wave $\tilde{z}_{i}$, $\tilde{z}_{j}$, its amplitude $\left|\boldsymbol{z}_{r}\right|$ and phase $\boldsymbol{\theta}_{r}$ can be calculated as follows:
\begin{equation}
\left|\boldsymbol{z}_{r}\right|=\sqrt{\left|\boldsymbol{z}_{i}\right|^{2}+\left|\boldsymbol{z}_{j}\right|^{2}+2\left|\boldsymbol{z}_{i}\right| \odot\left|\boldsymbol{z}_{j}\right| \odot \cos \left(\boldsymbol{\theta}_{j}-\boldsymbol{\theta}_{i}\right)},
    \label{eqa 2}  
\end{equation}
\begin{equation}
    \begin{aligned}
    \boldsymbol{\theta}_{r}=\boldsymbol{\theta}_{i}+\operatorname{\mathit{atan}} 2\left(\left|\boldsymbol{z}_{j}\right| \odot \sin \left(\boldsymbol{\theta}_{j}-\boldsymbol{\theta}_{i}\right),\right. \\
    \left.\left|\boldsymbol{z}_{i}\right|+\left|\boldsymbol{z}_{j}\right| \odot \cos \left(\boldsymbol{\theta}_{j}-\boldsymbol{\theta}_{i}\right)\right).
    \end{aligned}
\label{eaq 3}
\end{equation}

Inspired by WSiP~\cite{wsip}, we consider the target and surrounding agents as waves, and use the superposition of waves to dynamically model the interactions between agents. The phase discrepancy between agents $\left|\theta_{j}-\theta_{i}\right|$ significantly modulates the aggregated amplitude of the result $\boldsymbol{z}_{r}$, as depicted in Eq.~(\ref{eqa 2}) and illustrated in Figure~\ref{fig:wave}, thereby simulating the variable influence exerted by surrounding agents on the target agent.

\begin{figure*}[t]
\centering
\includegraphics[width=0.95\textwidth]{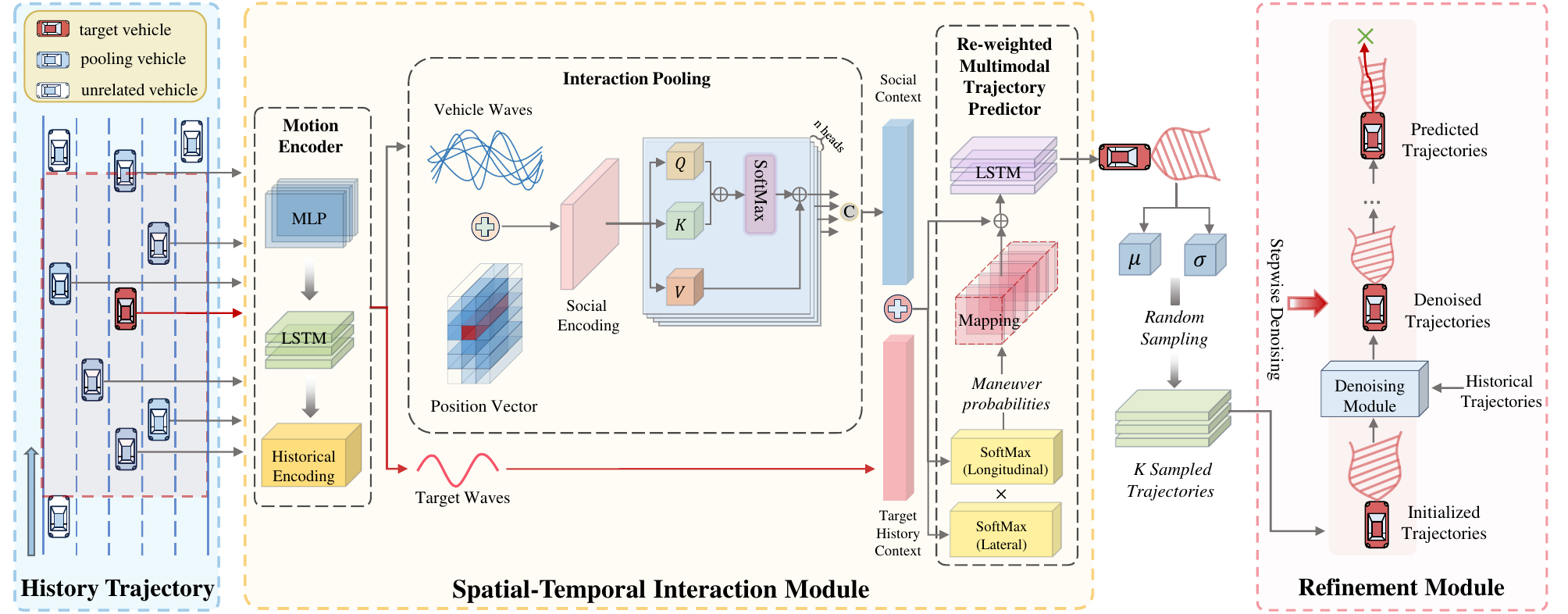} 
\caption{Framework of the proposed C2F-TP model, which contains a spatial-temporal interaction module and a refinement module. The spatial-temporal interaction module captures the dynamics and temporal correlations of inter-vehicle interactions and generates a multimodal trajectory distribution. A certain number of noisy trajectories are sampled from the multimodal trajectory distribution. The refinement module then denoises these noisy trajectories and finally generates accurate predicted trajectories.}
\label{framework}
\end{figure*}
\section{Methodology}
Figure~\ref{framework} shows the framework of C2F-TP, which is a coarse-to-fine denoising framework for uncertainty-aware vehicle trajectory prediction. C2F-TP contains a spatial-temporal interaction module and a refinement module. The former captures the dynamics and temporal correlation of inter-vehicle interactions and generates a multimodal future trajectory distribution for the target vehicle. Then we sample $k$ noisy trajectories from the multimodal trajectory distribution. Next, the refinement module denoises the $k$ trajectories based on a conditional denoising network and generates the final refined future trajectories.

\subsection{Spatial-Temporal Interaction Module}
In order to accurately capture the interactions between the target and surrounding agents, we propose the Spatial-Temporal Interaction Module to generate an initial future trajectory distribution. This module contains motion encoding, interaction pooling and re-weighted multimodal trajectory predictor. Next, we will describe the module in detail.

\subsubsection{Motion Encoding.}
Following the previous approach~\cite{cslstm, wsip}, we assume that the neighbors of the target agent (i.e., surrounding agents) are within $\pm90$ feet in the longitude direction and within four adjacent lanes centered on the target agent, as shown on the left side of Figure~\ref{framework}. Given the historical trajectories of the target agent $X_{tar}$ and the surrounding agents $\mathbb{X}_{\mathcal{N}}$, the module adopts a multi-layer perception (MLP) and a long short-term memory network (LSTM) to encode the high-level historical trajectory of the target agent $enc_\mathit{{tar}}$ and the surrounding agents $enc_\mathit{{nbrs}}$, respectively. 

\subsubsection{Interaction Pooling.}
Interactions between agents are highly dynamic and time-dependent. Simultaneously capturing the dynamic and temporal correlations of interactions between agents helps to more accurately model the future trajectories of target agents. 
Motivated by WSiP~\cite{wsip}, Interaction Pooling considers the agents in the road scene as waves, and uses the superposition of waves to dynamically model the interactions between agents. A multi-head self-attention is also adopted to capture the temporal correlations between social representations.
 
Inspired by Wave-MLP~\cite{wavemlp}, for any agent $a_i$, given the hidden state of the current timestamp $h_i$ of the agent, we use a plain fully connected layer (Plain-FC) to obtain the amplitude embedding $z_i$. Therefore, amplitude could represent the dynamics of the agent. Phase is used to modulate the aggregation of information from surrounding agents. We also use Plain-FC as follows to learn the phase embedding ${\theta}_{i}$ so that it can be dynamically adapted to the motion state of the agent,
\begin{equation}
\begin{split}
\boldsymbol{z}_{i}&=\mathit{Plain}\text{-}\mathit{FC}\left(\boldsymbol{h}_{i}, W^{z}\right),\\
\boldsymbol{\theta}_{i}&=\mathit{Plain}\text{-}\mathit{FC}\left(\boldsymbol{h}_{i}, W^{\theta}\right),
\end{split}
\end{equation}
where $W^{z}$ and $W^{\theta}$ are learnable weights. We design a Surrounding-FC module for aggregating different agent interaction information based on the token mixing module of Wave-MLP~\cite{wavemlp} as follows
\begin{equation}
\begin{split}
\tilde{\boldsymbol{o}}_{j}&=\mathit{Surrounding}\text{-}\mathit{FC}\left(\tilde{Z}, M_{\mathit{pos}}, W^{t}\right)_{j},\\
j&=1,2, \cdots, n,
\end{split}
\end{equation}
where $\tilde{\boldsymbol{Z}}=\left[\tilde{\boldsymbol{z}}_{1}, \tilde{\boldsymbol{z}}_{2}, \cdots, \tilde{\boldsymbol{z}}_{n}\right]$ denotes $n$ agent waves, $M_{pos}$ denotes the position mask, and $W^{t}$ is a learnable weight. The amplitude and phase information of different agents is aggregated by Surrounding-FC and the resulting  $\tilde{\boldsymbol{o}}_{j}$ is a complex-valued representation of the aggregated features. Then we obtain the real-valued output ${\boldsymbol{o}}_{j}$ by summing the real and imaginary parts of $\tilde{\boldsymbol{o}}_{j}$ with the weights~\cite{quantum} as follows
\begin{equation}
\begin{split}
    \boldsymbol{o}_{j}&=\sum_{k} W_{j k}^{t} \boldsymbol{z}_{k} \odot \cos \boldsymbol{\theta}_{k}+W_{j k}^{i} \boldsymbol{z}_{k} \odot \sin \boldsymbol{\theta}_{k}, \\
j&=1,2, \cdots, n,
\end{split}
\end{equation}
where $W^{t}$ and $W^{i}$ are learnable weights, $\odot$ is the element-wise multiplication and $j$ indicates the $j$-th output representation. In the above equation, the phase ${\theta}_{k}$ is dynamically adjusted according to the history states of various agents. Different agents with both the phase and amplitude information interact with each other through Surrounding-FC and obtain the social encoding.

The above social interaction feature is calculated for each timestamp $t$ separately without considering the temporal correlation. Thus, we use a multi-head self-attention to capture the relationship between the social interaction representations across different timestamps.
We denote the social encoding as $H$.

We take the sequence $H=\left(h^{t-T_h}, \cdots, h^{t-2}, h^{t-1}\right)$ as the input. Following self-attention mechanism, we calculate the query $(Q)$, key $(K)$, and value $(V)$ matrices based on the sequence $H$. 
The temporal self-attention is represented as
\begin{equation}
\mathcal{M}_{score}=\operatorname{\mathit{softmax}}\left(\left({Q}, K\right) / \sqrt{d_{k}}\right),
    \label{eqa }
\end{equation}
where $\mathcal{M}_{score} \in R^{T_h \times T_h}$ is the attentive score matrix whose entry $\mathcal{M}_{score}^{ij}$ measures the temporal correlation between timestamps $i$ and $j$. Then, the score matrix is used to aggregate temporal information from the corresponding values as follows
\begin{equation}
\textit{head}=\mathcal{M}_{\mathit{score}}{V}.
    \label{eqa 10}
\end{equation}
The output of the multiple self-attention is as follows
\begin{equation}
\textit{Multi-Head(Q,K,V)}=\textit{Concat}\left(\textit{head}_1,\textit{head}_2,\cdots, \textit{head}_k\right),
    \label{eqa 12}
\end{equation}
where $\textit{Multi-Head}(Q,K,V)$ are layer normalized to get the final social context $\tilde{H}=\left(\tilde{h}^{t-T_h}, \cdots, \tilde{h}^{t-2}, \tilde{h}^{t-1}\right)$ as shown in Figure 3.

\subsubsection{Re-weighted Multimodal Trajectory Predictor.}
To predict the multimodal trajectory of a vehicle, we propose the Re-weighted Multimodal Trajectory Predictor to generate the future trajectory modalities $\widehat{Y}$ with probability of $M$. 
Note that $M=\left\{m_{i} \mid i=1,2, \ldots, 6\right\}$ contains 6 modes, i.e., three lateral lane changing modes (turn left, turn right, and maintain lane), and two longitudinal speed shifting modes (braking and maintaining speed). We begin by merging the social context $\tilde{H}$ generated by the Interaction Pooling and the ego history context $enc_\mathit{{tar}}$ generated by the Motion Encoder to obtain the interaction context $\mathcal{C}$, which is then input into this module. 
As shown in Figure \ref{framework}, $\mathcal{C}$ is fed to two soft-max layers to output the lateral and longitudinal maneuver probabilities $P\left(m_{i} \mid \boldsymbol{X}\right)$, respectively. $X$ are the historical trajectories of agents in a scene.  

Considering the future trajectories of the different modes vary, previous methods simply combine modal representations with interaction context for prediction~\cite{cslstm}, which are unable to capture the relationship between modal representation pairs and interaction context at a fine-grained level. 
To address this problem, we design 6 mapping matrices to combine features at different historical timestamps in an adaptive manner inspired by~\cite{stddn}.

We use softmax to generate the corresponding 6 weight matrices as follows for each modality $u_i^{t_f}$, where $i$ denotes motion mode and $t_f$ denotes the future prediction horizon.
\begin{equation}
    u_i^{t_f} = \left(u^{-T_h, t_f}_{i}, \cdots, u^{-2, t_f}_{i}, u^{-1, t_f}_{i}\right),
    \label{}
\end{equation}
where $t_h = -T_h,\cdots, -2, -1$ denotes the historical prediction horizons.
Given the interaction context $\mathcal{C}=\left(c^{t-T_{h}}, \ldots, c^{t-2}, c^{t-1}\right)$, we use six $u^i_{t_f}$ for each mode to adaptively combine $\mathcal{C}_{\text{tar}}$ separately as follows
\begin{equation}
    \mathcal{V} ^{t_f}=\sum_{t=-T_h}^{-1} c^{t} u^{t,t_f}_{i}.
    \label{eqa 11}
\end{equation}

Finally, the weighted modal mapping vectors $\mathcal{V} ^{t_f}$ are fed into an LSTM layer along with the interaction context $\mathcal{C}$, which can output the parameters of a bivariate Gaussian distribution $\widehat{Y}$ of the target agent in each maneuver modality. The predicted locations $\widehat{Y}^t_i$ at time step $t$ are as follows
\begin{equation}
    \widehat{Y}^t_i \sim \mathcal{N} \left ({\mu}^t_i, {\sigma }^t_i, {\rho }^t_i\right ) ,
\end{equation}
where ${\mu}^t_i$ and ${\sigma }^t_i$ are the means and variances of future locations respectively, and ${\rho }^t_i$ is the correlation coefficient. We denote ${\Omega}$ as parameters of Gaussian distribution. The posterior probability of the future trajectories are as follows,
\begin{equation}
P(\boldsymbol{Y} \mid \boldsymbol{X})=\sum_{i} P\left(m_{i} \mid \boldsymbol{X}\right) P_{\boldsymbol{\Omega}}\left(\boldsymbol{Y} \mid m_{i}, \boldsymbol{X}\right).
    \label{eqa 12}
\end{equation}

\begin{table*}
\small 
\centering
\setlength{\tabcolsep}{3.4mm}
\begin{tabular}{ccccccccc}
\toprule 
\multirow{2}{*}{Horizon} & \multicolumn{8}{c}{RMSE (NGSIM/HighD)}\\
\specialrule{0em}{1pt}{1pt}
\cline{2-9}
\specialrule{0em}{1pt}{1pt}
&V-LSTM    & S-LSTM    & CS-LSTM   & S-GAN     & PiP-no Plan & STDAN     & WSiP      & C2F-TP (C) \\
\specialrule{0em}{1pt}{1pt}
\midrule 
\specialrule{0em}{1pt}{1pt}
1s& 0.68/0.22 & 0.59/0.21 & 0.58/0.24 & 0.57/0.30 & 0.57/0.21   & \underline{0.42}/\underline{0.15} & 0.56/0.20 & \textbf{0.32/0.11} \\
\specialrule{0em}{1pt}{1pt}
2s& 1.66/0.65 & 1.29/0.65 & 1.27/0.68 & 1.32/0.78 & 1.24/0.62   & \underline{1.01}/\underline{0.45} & 1.23/0.60 & \textbf{0.92/0.41} \\
\specialrule{0em}{1pt}{1pt}
3s& 2.96/1.32 & 2.13/1.31 & 2.11/1.26 & 2.22/1.46 & 2.05/1.26   & \underline{1.69}/\underline{0.94} & 2.05/1.21 & \textbf{1.62/0.92} \\
\specialrule{0em}{1pt}{1pt}
4s& 4.56/2.22 & 3.21/2.16 & 3.19/2.15 & 3.26/2.34 & 3.07/2.14   & \underline{2.56}/\underline{1.68} & 3.08/2.07 & \textbf{2.44/1.64} \\
\specialrule{0em}{1pt}{1pt}
5s& 5.44/3.43 & 4.55/3.29 & 4.53/3.31 & 4.41/3.41 & 4.34/3.27   & \underline{3.67}/\textbf{2.58} & 4.34/3.14 & \textbf{3.45}/\underline{2.60} \\
\specialrule{0em}{1pt}{1pt}
\midrule 
Average& 3.06/1.57 & 2.35/1.52 &2.34/1.53 &2.36/1.66 &2.25/1.50 &\underline{1.87}/\underline{1.16} &2.25/1.44 &\textbf{1.75}/\textbf{1.14} \\
\bottomrule 
\end{tabular}
\caption{Results of the comparison between the initial trajectory distributions obtained in the first stage C2F-TP (C) and baselines on the NGSIM and highD datasets. We report the RMSE for 5s prediction horizon. Bold indicates the best result, and underline indicates the second-best result.}
\label{fig:tabel1}
\end{table*}
\subsection{Refinement Module}
The highly dynamic interactions between vehicles make the trajectory uncertainty transmit among vehicles, increasing the challenge of accurately predicting future trajectories~\cite{multi}. To address this issue, we sample out $k$ noisy trajectories $\widehat{Y}_{k}$ based on the future trajectory distribution $\widehat{Y}$ from the first stage to model the uncertainty. In the inverse diffusion process, the refinement module reduces the uncertainty of the trajectories by progressively denoising them to generate more accurate and reliable future trajectories.

Specifically, the refinement module is a conditional denoising model that denoises the trajectory $\widehat{Y}_{k}$ based on historical trajectories ($X_{tar}$ and $\mathbb{X}_{\mathcal{N}}$) and generates optimized future trajectories by gradual denoising steps. This module contains a transformer-based context encoder $f_{context}(\cdot)$ to learn a social-temporal embedding and a noise estimation module $f_{\epsilon}(\cdot)$ to estimate the noise to reduce. The $t$-th denoising step works as follows
\begin{equation}
\begin{split}
    \chi &= f_{context} \left(X_{tar}, \mathbb{X}_{\mathcal{N}}\right), \\
    \epsilon_{\theta}^{t}&=f_{\boldsymbol{\epsilon}}\left(\widehat{\mathbf{Y}}_{k}^{t+1}, \chi, t+1\right),
    \end{split}
\label{eqa 13}
\end{equation}
\begin{equation}
    \widehat{\mathbf{Y}}_{k}^{t}=\frac{1}{\sqrt{\alpha_{t}}}\left(\widehat{\mathbf{Y}}_{k}^{t+1}-\frac{1-\alpha_{t}}{\sqrt{1-\bar{\alpha}_{t}}} \epsilon_{\theta}^{t}\right)+\sqrt{1-\alpha_{t}} \mathbf{z},
    \label{eqa 14}
\end{equation}
where $\alpha_{t}$ and $\bar{\alpha}_{t}=\prod_{i=1}^{t} \alpha_{i}$ are the parameters in the diffusion process and $\mathrm{z} \sim \mathcal{N}(\mathbf{z} ; \mathbf{0}, \mathrm{I})$ is a noise. $f_{\epsilon}(\cdot)$ estimates the noise $\epsilon_{\theta}^{t}$ in the noisy trajectory $\widehat{Y}_{k}$ implemented by multi-layer perceptions with the historical trajectories. Eq.~(\ref{eqa 14}) is a standard denoising process.  
This approach significantly reduces the uncertainty in trajectory prediction and further improves the accuracy of predicted trajectories.

\begin{table*}[!h]
\small
\centering
\setlength{\tabcolsep}{3.4mm}
\begin{tabular}{ccccccccc}
\toprule 
\multirow{2}{*}{Dataset} & \multirow{2}{*}{Horizon} & \multicolumn{7}{c}{Metric (ADE/FDE) (m)} \\
\specialrule{0em}{1pt}{1pt}
\cline{3-9}
\specialrule{0em}{1pt}{1pt}
                         & & V-LSTM    & S-LSTM    & CS-LSTM   & WSiP      & LED       & C2F-TP (C) & C2F-TP \\
\specialrule{0em}{1pt}{1pt}
\midrule 
\specialrule{0em}{1pt}{1pt}
\multirow{7}{*}{NGSIM}   & 1s & 0.22/0.42 & 0.22/0.40 &\underline{0.21}/\underline{0.40} & 0.30/0.42 & 0.40/0.41 & 0.26/0.43 & \textbf{0.20/0.34}   \\
\specialrule{0em}{1pt}{1pt}
                         & 2s & 0.53/1.16 & 0.52/1.23 & 0.51/1.10 & 0.63/1.28 & 0.64/\textbf{0.72} & \underline{0.50}/0.95 & \textbf{0.47}/\underline{0.95}   \\
                         \specialrule{0em}{1pt}{1pt}
                         & 3s & 0.93/2.14 & 0.89/2.08 & \underline{0.88}/1.81 & 1.03/2.22 & 0.91/\textbf{1.10} & 0.78/1.56 & \textbf{0.78}/\underline{1.47}   \\
                         \specialrule{0em}{1pt}{1pt}
                         & 4s & 1.40/3.35 & 1.42/3.30 & 1.25/2.85 & 1.49/3.37 & 1.20/\underline{1.39} & \underline{1.09}/2.25 & \textbf{1.08/1.35}   \\
                         \specialrule{0em}{1pt}{1pt}
                         & 5s & 1.94/4.76 & 1.89/4.37 & 1.69/4.08 & 2.03/4.71 & 1.52/\underline{2.02} & \textbf{1.41}/3.09 & \underline{1.45}/\textbf{1.36}   \\
                         \specialrule{0em}{1pt}{1pt}
\cline{2-9}
\specialrule{0em}{1pt}{1pt}
                     & Average &1.00/2.37 & 0.99/2.28 & 0.91/2.05 & 1.10/2.40 & 0.93/\underline{1.13} & \underline{0.81}/1.66 & \textbf{0.79}/\textbf{1.09} \\
                     \specialrule{0em}{1pt}{1pt}
\cline{1-9}
\specialrule{0em}{1pt}{1pt}
\multirow{7}{*}{HighD}   & 1s & 0.29/0.48 & 0.23/0.37 & 0.21/0.35 & \underline{0.15}/\underline{0.23} & 0.50/0.81 & \underline{0.15}/0.26 & \textbf{0.14/0.20}   \\
\specialrule{0em}{1pt}{1pt}
                         & 2s & 0.53/0.98 & 0.41/0.76 & 0.39/0.72 & 0.28/0.60 & 0.94/1.61 & \underline{0.27}/\underline{0.49} & \textbf{0.23/0.32}   \\
                         \specialrule{0em}{1pt}{1pt}
                         & 3s & 0.78/1.53 & 0.60/1.20 & 0.57/1.14 & 0.45/1.09 & 1.41/2.39 & \underline{0.40}/\underline{0.88} & \textbf{0.33/0.56}   \\
                         \specialrule{0em}{1pt}{1pt}
                         & 4s & 1.04/2.14 & 0.81/1.68 & 0.77/1.61 & 0.69/2.01 & 1.91/3.19 & \underline{0.57}/\underline{1.47} & \textbf{0.44/0.53}   \\
                         \specialrule{0em}{1pt}{1pt}
                         & 5s & 1.32/2.81 & 1.03/2.22 & 0.99/2.14 & 1.00/3.19 & 2.46/3.98 & \underline{0.78}/\underline{2.24} & \textbf{0.59/0.53}   \\
                         \specialrule{0em}{1pt}{1pt}
\cline{2-9}
\specialrule{0em}{1pt}{1pt}
                        &Average &0.79/1.59 &0.62/1.25 &0.59/1.19 &0.51/1.42 &1.44/2.40 &\underline{0.43}/\underline{1.07} &\textbf{0.35}/\textbf{0.43}\\
                        \specialrule{0em}{1pt}{1pt}
\bottomrule 
\end{tabular}
\caption{Results of the comparison between baselines and our method (C2F-TP) on the NGSIM and highD datasets. We report two error metrics ADE and FDE for 5s prediction horizon. Bold indicates the best result, and underline indicates the second-best result.}
\label{tab:table2}
\end{table*}
\section{Experiment}
\subsection{Dataset}
The experiments are conducted on two datasets \textbf{NGSIM}~\cite{cslstm} and \textbf{highD}~\cite{highd} that are widely adopted in trajectory prediction evaluation. The NGSIM dataset contains trajectories of real freeway traffic captured at 10 Hz over a time span of 45 minutes in 2015. The highD dataset consists of trajectories of 110,000 vehicles recorded at 25 Hz, which are collected at a segment of two-way roads around Cologne in Germany from 2017 to 2018. The dataset details are given in the associated code repository.

\subsection{Evaluation Metrics}
We employ Root Mean Square Error (RMSE), a widely used metric in trajectory prediction, to evaluate the spatial-temporal interaction stage as follows
\begin{equation}
    \mathit{RMSE}(t)=\sqrt{\frac{1}{N} {\textstyle \sum_{i=1}^{N}}\left(Y_{i}^{t}-\hat{Y}_{i}^{t}\right)^{2}},
    \label{eqa 16}
\end{equation}
where $N$ is the total number of test instances, $Y_{i}^{t}$ and $\hat{Y}_{i}^{t}$ are the ground-truth and predicted coordinates of agent $a_i$ at time step $t$, respectively. 

Additionally, at the end of the refinement stage, we adopt the Average Displacement Error (ADE) and Final Displacement Error (FDE) as follows, which are commonly used in denoising models, to assess the accuracy of the trajectories generated by our method.
\begin{equation}
\begin{split}
    \mathit{ADE}&=\frac{{\textstyle \sum_{i=1}^{N}} {\textstyle \sum_{t=1}^{T_f}}\left\|Y_{i}^{t}-\hat{Y}_{i}^{t}\right\|_{2}}{N \times T_{f}},\\
    \mathit{FDE}&=\frac{{\textstyle \sum_{i=1}^{N}} \left\|Y_{i}^{t}-\hat{Y}_{i}^{t}\right\|_{2}}{N}.
\end{split}
\end{equation}

\subsection{Baseline}
In the spatial-temporal interaction stage, we compare the generated trajectory distribution with six baselines: \emph{V-LSTM}~\cite{lstm}, \emph{S-LSTM}~\cite{Sociallstm}, \emph{CS-LSTM}~\cite{cslstm}, \emph{PiP-noPlan}~\cite{pip}, \emph{STDAN}~\cite{stddn}, and \emph{WSiP}~\cite{wsip}. We also compare the predicted final trajectory by the refinement module in the second stage with five state-of-the-art baselines: \emph{V-LSTM}~\cite{lstm}, \emph{S-LSTM}~\cite{Sociallstm}, \emph{CS-LSTM}~\cite{cslstm}, \emph{WSiP}~\cite{wsip} and \emph{LED}~\cite{led}.
The detailed introduction of baselines is given in the associated code repository.

To gain insight into the effect of the key components of C2F-TP, we compare C2F-TP with its three variants.
\begin{itemize}
\item \emph{\textbf{w/o\_IP}}: C2F-TP without the Interaction Pooling. 
\item \emph{\textbf{w/o\_RWMTP}}: C2F-TP without the Re-weighted Multimodal Trajectory Predictor.
\item \emph{\textbf{C2F-TP(C)}}: C2F-TP without the Refinement module. 
\end{itemize}

\subsection{Implementation Details}
We implement our model with the Pytorch framework on a GPU server with NVIDIA 3090 GPU. The parameters in the model are set as follows. We employ a $13\times 5$ grid, which is defined around the target vehicle, where each column corresponds to a single lane, and the rows are separated by a distance of 15 feet. The hidden features of MLP layers are set to 32 with ReLu as the activation function. 
To train a coarse-to-fine framework, we consider a two-stage training strategy, where the first stage trains a denoising module and the second stage focuses on training a spatial-temporal interaction module. 
The details of the two-stage prediction process are given in the associated code repository. Each trajectory is split into segments over a horizon (i.e., 8s), which contains the past (3s) and future (5s) positions at 5Hz. We split the dataset into training, validation, and testing sets with a splitting ratio of $7:2:1$. Note that the baseline methods are set based on their original papers and the accompanying code. 


\subsection{Overall Performance Comparison}
To study the effectiveness of the C2F-TP, we compare it with 6 baselines for the spatial-temporal interaction stage (C2F-TP (C)) and with 5 baselines for the refinement stage (C2F-TP), respectively. Based on the result shown in the Table 1 and Table 2, we can have the following observations.

\noindent\textbf{\textit{Spatial-Temporal Interaction Stage.}} Table~\ref{fig:tabel1} shows the performance comparison among different methods on the two datasets for the spatial-temporal interaction stage. C2F-TP (C) achieves the best results among all baselines in most cases, performing better than the best among the baselines by up to 24\% and 26.7\% on NGSIM and highD, respectively in terms of RMSE. As a popular trajectory prediction model, STDAN performs the best among all the baselines in most cases due to its powerful ability to capture spatial and temporal correlations. Averagely, C2F-TP (C) performs better than STDAN by up to 6.4\% and 2.1\% on NGSIM and highD, respectively. This is because of the dynamic modeling of vehicle interactions with wave superposition, which facilitates trajectory prediction.

\begin{figure*}[h!]
    \centering
    \subfigure[ADE on NGSIM]{\includegraphics[width=0.23\linewidth]{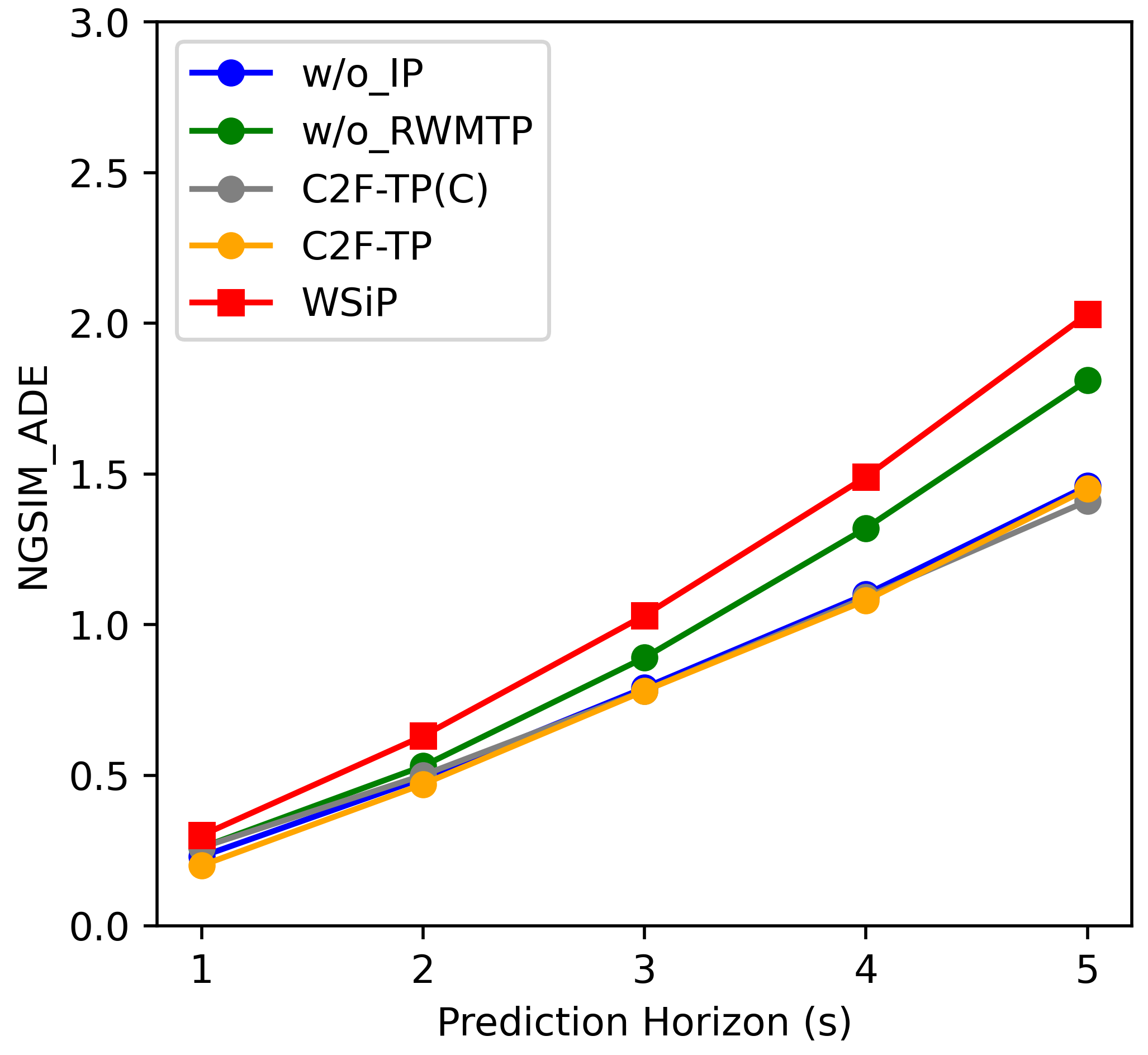}}
    \subfigure[FDE on NGSIM]{\includegraphics[width=0.23\linewidth]{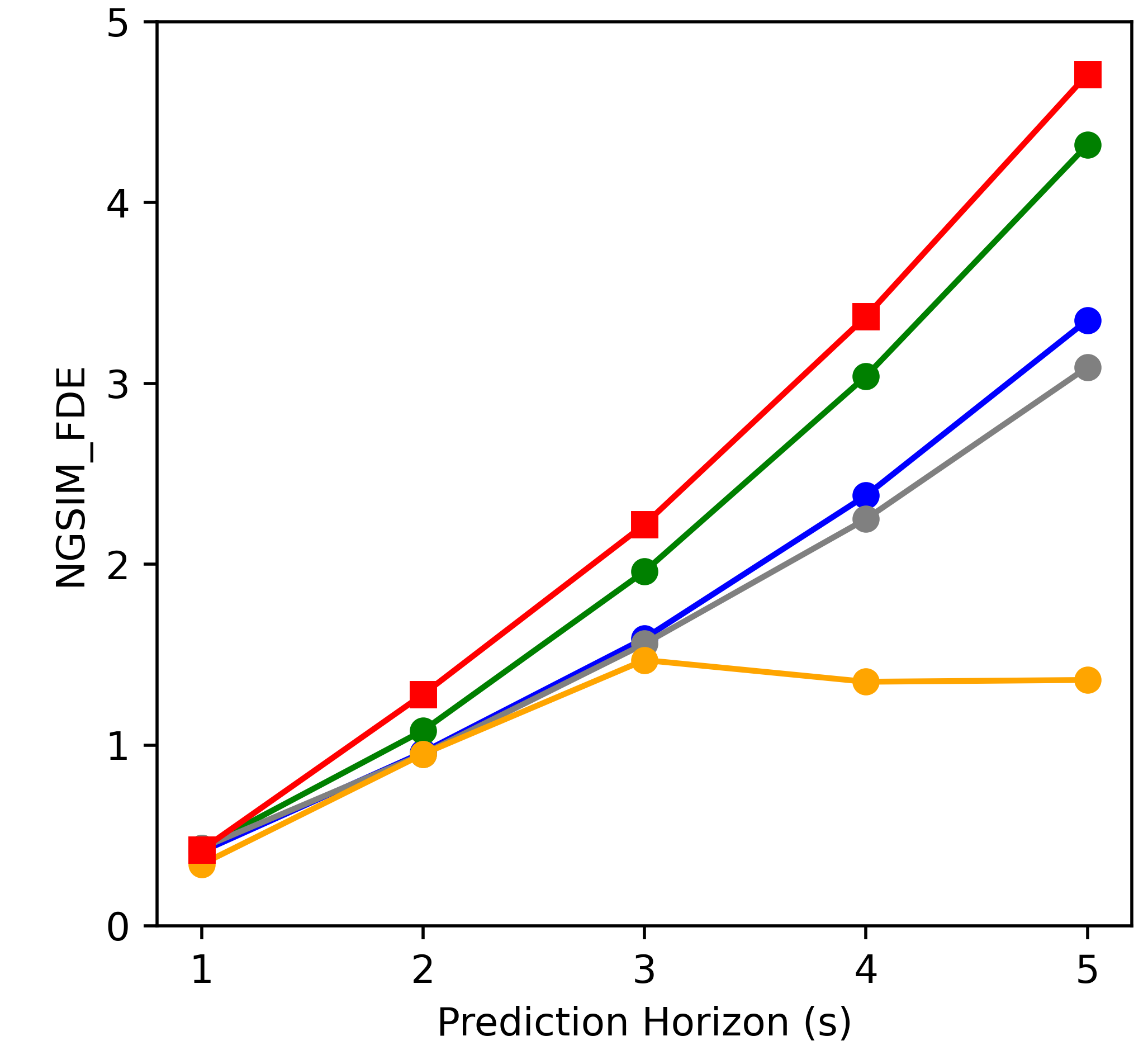}}
    \subfigure[ADE on highD]{\includegraphics[width=0.23\linewidth]{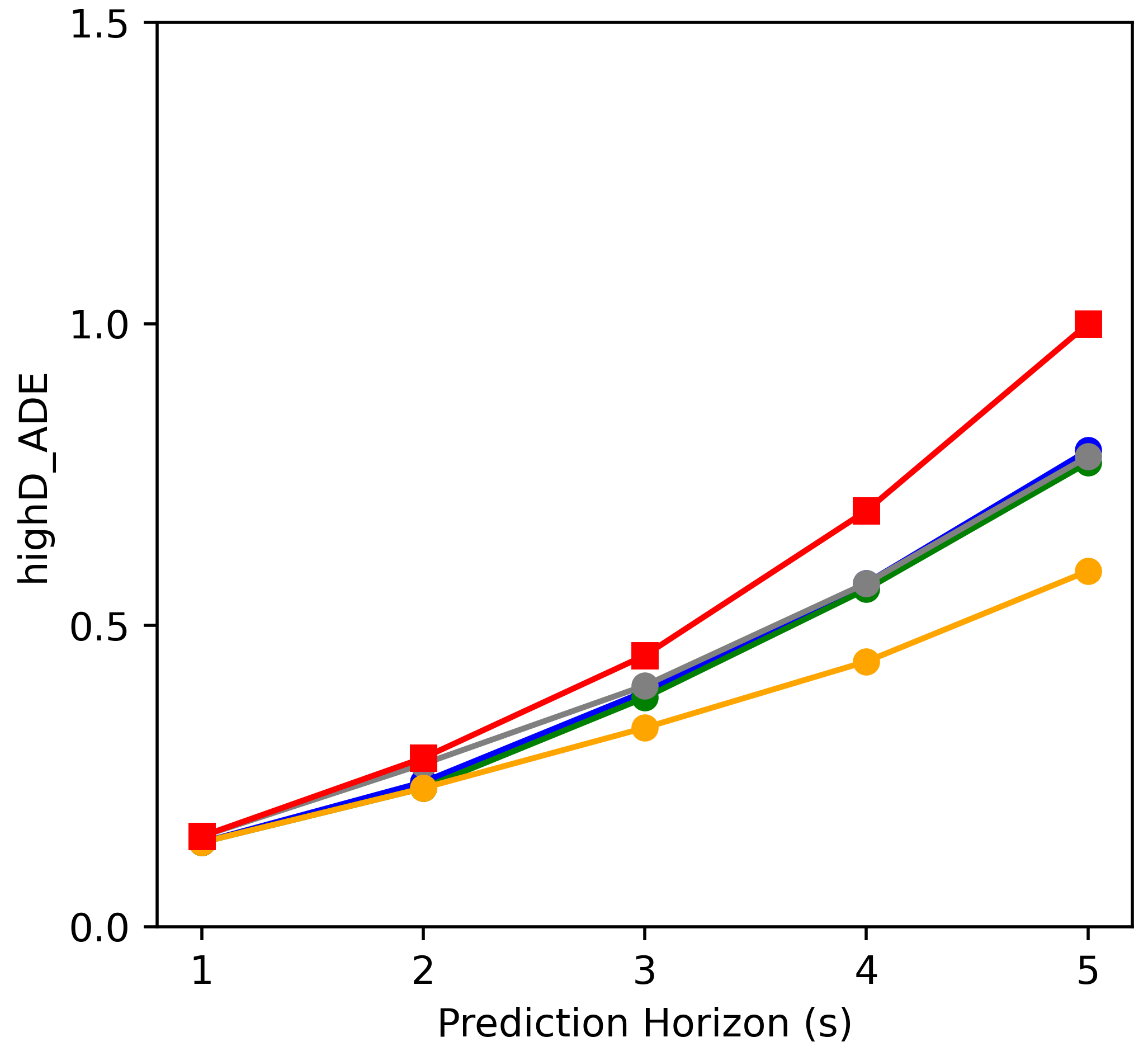}}
    \subfigure[FDE on highD]{\includegraphics[width=0.23\linewidth]{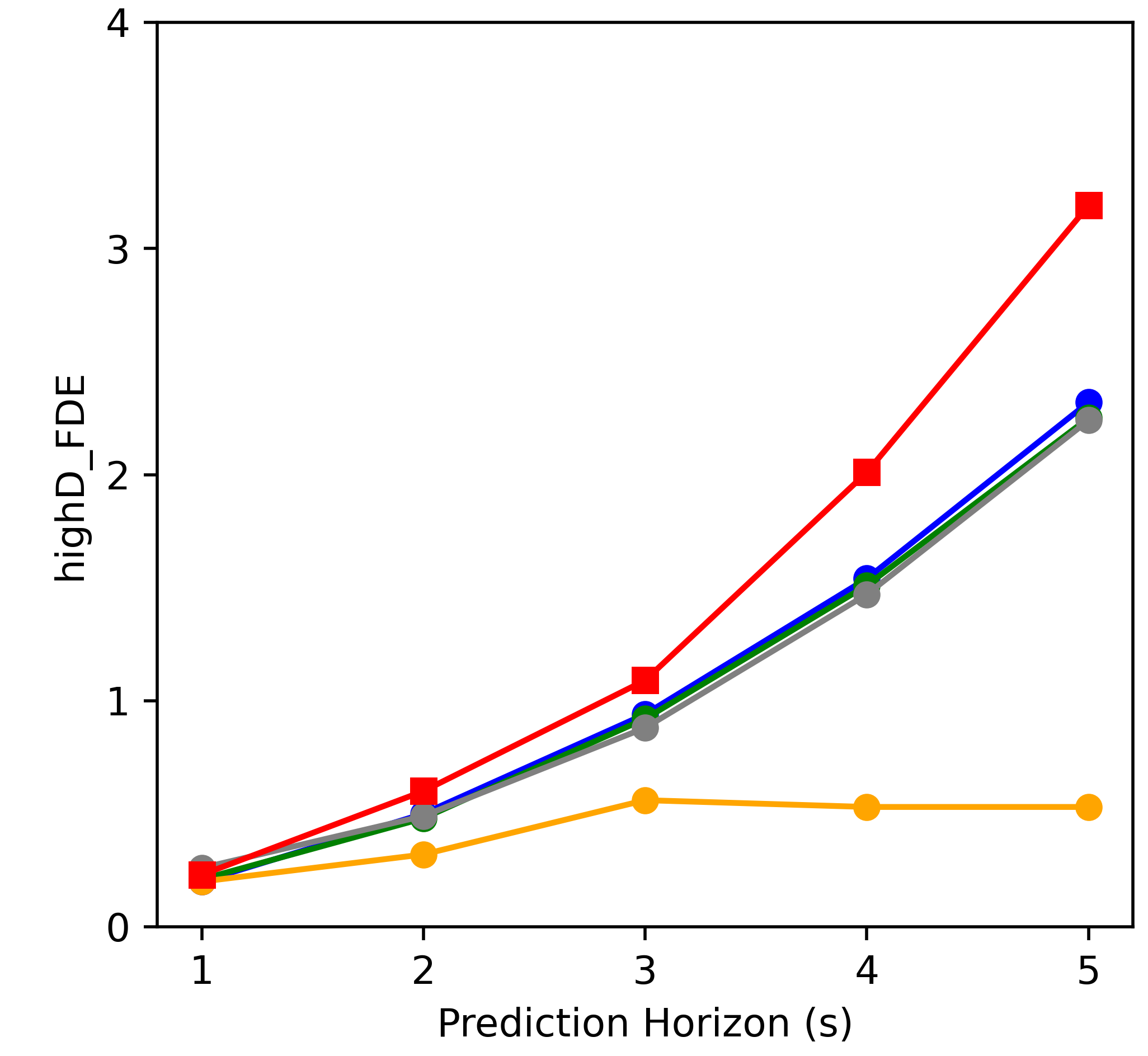}}
    \caption{Comparison between C2F-TP and its variants.}
    \label{fig:ablation}
\end{figure*}
\begin{figure*}[!tbp]
    \centering
    \subfigure[Keeping straight]{\includegraphics[width=0.32\textwidth]{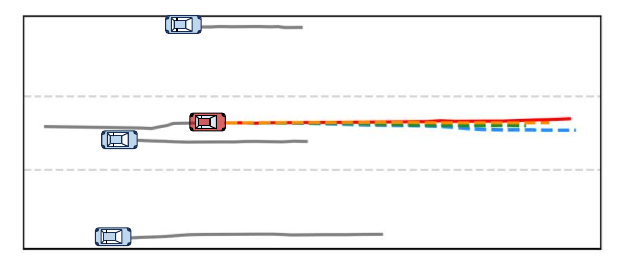}\label{casea}}
     \subfigure[Merging to the left lane]{\includegraphics[width=0.32\textwidth]{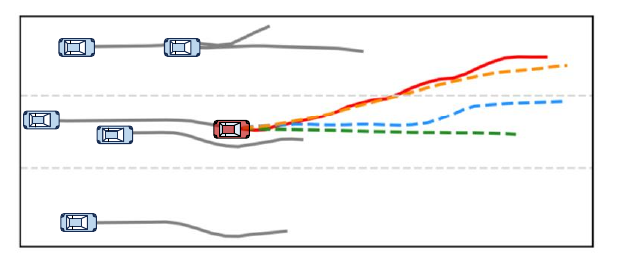}\label{caseb}}
      \subfigure[Merging to the right lane]{\includegraphics[width=0.32\textwidth]{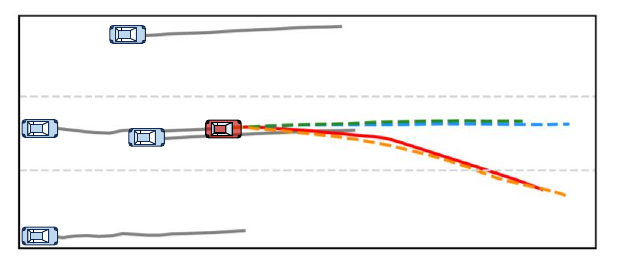}\label{casec}}
      \subfigure{\includegraphics[width=0.9\textwidth]{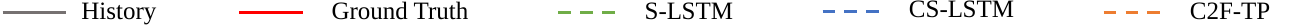}}
      \caption{Visualisation of S-LSTM, CS-LSTM and C2F-TP for three driving scenarios.}
      \label{casestudy}
\end{figure*}

\noindent \textit{\textbf{Refinement Stage.}} The sampled trajectories from the spatial-temporal interaction stage are fed into the refinement module.
Table~\ref{tab:table2} reports the ADE and FDE results on both datasets when the prediction horizon is set to 5s. We observe that C2F-TP   outperforms baselines on NGSIM in most cases, while it achieves the best performance on highD. More specifically, C2F-TP outperforms the best among the baselines by 6.7\%-40.4\% and 13.0\%-75.2\% for ADE and FDE on highD dataset, respectively. In addition, C2F-TP brings significant performance improvement on long-term trajectory prediction, especially on highD. This is because the refinement module captures the randomness and uncertainty of the trajectories, thereby increasing the accuracy and stability of the trajectory prediction through the gradual denoising process. Additionally, C2F-TP outperforms C2F-TP (C) by up to 76\% due to the consideration of uncertainty, which generates more realistic predicted trajectories.
We report the effect of denoising steps in the associated code repository.

\subsection{Ablation Study}
To assess whether the components in C2F-TP all contribute to the performance of trajectory prediction, we compare C2F-TP with its three variants. The results are shown in Figure~\ref{fig:ablation}. We have the following observations from the result.
\begin{itemize}
    \item Regardless of the datasets, C2F-TP always performs better than its counterparts without the interaction pooling, the re-weighted multi-trajectory predictor, and the refinement module. This shows these three components are all useful for effective trajectory prediction.
    \item w/o\_RWMTP performs worst among all variants, which shows the importance of the re-weighted multimodal trajectory predictor. It offers evidence that the re-weighted multimodal trajectory predictor can learn comprehensive driving behaviors. C2F-TP and its variants consistently outperform WSiP, suggesting that modeling the time-dependent performance of interactions between intelligences captures a more complete characterization of the interactions and thus predicts more accurate trajectories.
    \item With an increase in prediction horizon, C2F-TP increasingly outperforms its variants due to its capabilities in long-term trajectory prediction.
\end{itemize}

\subsection{Case Study}
We visualize the trajectory prediction results for several cases in Figure~\ref{casestudy} to intuitively show the effectiveness of C2F-TP. Three driving scenarios are selected including target keeping straight, merging to the left lane, and merging to the right lane. 
As shown in Figure~\ref{casea}, S-LSTM, CS-LSTM, and C2F-TP all achieve acceptable performance when the target keeps going straight. C2F-TP can trace the right trajectory more accurately. Figure~\ref{caseb} shows that the target merges to the left lane. Both CS-LSTM and C2F-TP make correct direction predictions, but the predicted trajectory of C2F-TP is closer to the ground truth. Figure~\ref{casec} shows the target merges to the right lane. The predictions of S-LSTM and CS-LSTM keep going straight, while C2F-TP predicts that the target will merge into the right lane. This is because of the re-weighted trajectory predictor, which considers multi-modal modeling enabling possible multiple trajectory prediction.

\section{Conclusion}
This paper proposed a coarse-to-fine trajectory prediction framework C2F-TP based on a novel two-stage generation process for trajectory prediction by considering the dynamic and temporal interactions across vehicles. Specifically, the spatial-temporal interaction stage employed an interaction pooling module to model the driving dynamics and time-dependence of inter-vehicle interactions. A re-weighted multimodal trajectory predictor was proposed to fuse specific interaction features based on specific modalities, facilitating the learning on the multimodal trajectory distributions. Then, a certain number of trajectories were sampled from the learned trajectory distribution, which was then fed into a refinement module based on a denoising diffusion model. The refinement module aimed to reduce the data uncertainty and generate the accurate trajectory predictions. An empirical study on two real datasets offered evidence of the effectiveness of C2F-TP.

\section{Acknowledgments}
This research was funded by the National Natural Science Foundation of China (No.62172443 and 62472446), Hunan Provincial Natural Science Foundation of China (No.2022JJ30053) and the High Performance Computing Center of Central South University.

\bibliography{aaai25}

\section{Appendix}
\subsection{Datasets}
The experiments are carried out on two real-world public trajectory datasets: NGSIM and highD.
\begin{itemize}
    \item \emph{NGSIM:} The NGSIM dataset contains detailed vehicle trajectory information such as vehicle’s coordinates, velocity, etc., on eastbound I-80 in the San Francisco Bay area and southbound US 101 in Los Angeles. This dataset was collected by the U.S. Department of Transportation in the year of 2015. This dataset consists of real highway driving scenarios recorded by multiple overhead cameras at 10Hz.
    \item \emph{highD:} The highD dataset is collected at a segment of about 420m of two-way roads around Cologne in German from drone video recordings at 25 Hz in the year of 2017 and 2018. It consists of 110 500 vehicles including cars and trucks and a total driven distance of 44 500 km. The dataset includes four files for each recording: an aerial 
    shot of the specific highway area and three CSV files, containing information about the site, the vehicles and the extracted trajectories.
\end{itemize}
The details of the dataset are shown in Table 3. We split all the trajectories contained in NGSIM and highD separately, in which 70\% are used for training with 20\% and 10\% for testing and evaluation. We split each of the trajectories
into 8s segments consisting of 3s of past and 5s of future
trajectories.

\subsection{Baselines}
We compare the proposed C2F-TP with the following baselines.
\begin{itemize}
\item \emph{V-LSTM}: Vanilla LSTM (V-LSTM) uses a single LSTM to encode historical trajectories of the target vehicle.
\item \emph{S-LSTM}: Social LSTM (S-LSTM) uses fully connected layers and generates the uni-modal distribution of the future locations.
\item \emph{CS-LSTM}: Convolutional Social LSTM (CS-LSTM) uses convolutional social pooling and generates multi-modal trajectory predictions.
\item \emph{PiP-noPlan}: PiP uses convolutional social pooling and a fully convolutional network to generate multi-modal trajectory predictions. We remove the planning coupled module (PiP-noPlan) as the future motions of the controllable ego vehicle is unavailable in real-world driving scenarios. 
\item \emph{STDAN}: STDAN captures multi-modal driving behaviors by hierarchically modeling motion states, social interactions, and temporal correlations in vehicle trajectories.
\item \emph{WSiP}: WSiP uses wave pooling and also generates multi-modal trajectory predictions.
\item \emph{LED}: LED leverages a trainable leapfrog initializer to directly learn an expressive multi-modal distribution of future trajectories, which skips a large number of denoising steps.
\end{itemize}

\begin{table}[t]
\small
\centering
\renewcommand{\arraystretch}{1} 
\begin{tabular}{ccccc}
\toprule 
\specialrule{0em}{1pt}{1pt}
Dataset & Agents   & Scene   & \begin{tabular}[c]{@{}c@{}}Duration \\ and \\ tracking quantity\end{tabular} & Data type \\
\specialrule{0em}{1pt}{1pt}
\hline
\specialrule{0em}{1pt}{1pt}
NGSIM   & vehicles & highway & \begin{tabular}[c]{@{}c@{}}90 min recording \\of two highways\end{tabular}  & \begin{tabular}[c]{@{}c@{}}trajectories,\\ lane\end{tabular} \\ 
\specialrule{0em}{1pt}{1pt}
\hline
\specialrule{0em}{1pt}{1pt}
highD   & vehicles & highway & \begin{tabular}[c]{@{}c@{}}110500 vehicles, \\ 44500 driven \\kilometers, \\147 driven hours\end{tabular} & \begin{tabular}[c]{@{}c@{}}trajectories,\\ lane\end{tabular} \\ 
\specialrule{0em}{1pt}{1pt}
\bottomrule 
\end{tabular}
\caption{Statistics of datasets.}
\end{table}

\subsection{Training Details}
In this section, we first show the objective function of spatial-temporal interaction module and C2F-TP as follows. 
\subsubsection{Spatial-Temporal Interaction Module.}
In the spatial-temporal interaction module, we aim to output a multimodal distribution of future trajectories. The objective is to minimize the negative log-likelihood loss $\mathcal{L}$ of the true trajectory under the class of maneuvers with maximum probability $m_{max}$ of the target. The loss function is defined as follows.
\begin{equation}
    \mathcal{L}=-\log \left(\mathrm{P}_{\Theta}\left(\mathbf{Y} \mid m_{\text {max}}, \mathcal{X}\right) \mathrm{P}\left(m_{\text {max}} \mid \mathcal{X}\right)\right),
\end{equation}
We train the model using Adam for 100 epochs with an initial learning rate of 0.001 and decay by 0.6 every 16 epochs.

\subsubsection{C2F-TP.}
Due to the large number of model parameters, in order to reduce the complexity of training, we use a two-stage training strategy to train the C2F-TP, where the first stage trains a refinement module based on a conditional denoising model, and the second stage focuses on a spatial-temporal interaction module. 

In the first stage, we train the refinement module based on a standard training schedule of a diffusion models through the noise estimation loss.
\begin{equation}
\mathcal{L}_{\mathrm{NE}}=\left\|\boldsymbol{\epsilon}-f_{\boldsymbol{\epsilon}}\left({\mathbf{Y}}^{t+1}, f_{context}\left({X_{tar}}, \mathbb{X}_{\mathcal{N}}\right), t+1 \right)\right\|_{2},
\end{equation}
where $t=1,2,\cdots,T$ ($T$ is the diffusion steps), $\epsilon \sim \mathcal{N} (\mathbf{\epsilon } ;\mathbf{0},\mathbf{I}) $ and $\mathbf{Y}^{t+1} = \sqrt{\bar{\alpha}_{t}}Y^0 + \sqrt{1-\bar{\alpha}_{t}}\epsilon $ is the diffused trajectory. We then backpropagate the loss and train the parameters in the context encoder $f_{context}(\cdot )$ and the noise estimation module $f_{\boldsymbol{\epsilon}}(\cdot )$.

In the second stage, we freeze the parameters of the refinement module and train the spatial-temporal interaction module. The loss function is defined as follows
\begin{equation}
    \mathcal{L} = \min _{k}\left\|\mathbf{Y}-\widehat{\mathbf{Y}}_{k}\right\|_{2}.
\end{equation}
We train the model using Adam for 20 epochs with an initial learning rate of 0.001 and decay by 0.5 every 6 epochs.

Next, we provide Algorithm~\ref{algorithm} to show the inference process of C2F-TP.
Line 1 indicates that the spatial-temporal interaction module outputs the distribution of future trajectories for different modalities. 
Line 2 indicates that $K$ noisy trajectories are sampled from the distribution of the maximum probability modality. Lines 3-7 shows the $\tau$-step denoising process of the $K$ sampled trajectories by the refinement module.
\begin{algorithm}[h!]
    \caption{C2F-TP in Inference}
    \label{algorithm}
    \begin{algorithmic}[1] 
        \Require Historical trajectories of target $\mathrm{X_{tar}}$ and its surrounding neighbors $\mathbb{X}_{\mathcal{N}}$
        \Ensure Future trajectory of target $\mathrm{Y_{tar}}$
        \State $\widehat{Y}_i \sim \mathcal{N} \left ({\mu}_i, {\sigma}_i, {\rho}_i \right)$ = \emph{Interaction} $\left(\mathrm{X_{tar}}, \mathbb{X}_{\mathcal{N}}\right)$  
        \State $\widehat{Y}_{k}^{\tau}$ $\sim$ \emph{Sample from } $\widehat{Y}_{max}$ \emph{with the maximum modal probability}
        \For {$t = \tau - 1 \to 0$} 
            \State $\widehat{Y}_{k}^t $ = \emph{Refinement} $\left(\widehat{Y}_{k}^{\tau}, \mathrm{X_{tar}}, \mathbb{X}_{\mathcal{N}}\right) $
        \EndFor
        \State $\widehat{\mathcal{Y}} = \left \{\widehat{Y}_{1}^{0}, \widehat{Y}_{2}^{0}, \cdots, \widehat{Y}_{k}^{0}  \right \} $
        \State \Return $\widehat{\mathcal{Y}}$
    \end{algorithmic}
\end{algorithm}

\subsection{Hyper-Parametric Analysis}
The traditional reverse diffusion process reduces noise in trajectories by gradually denoising, often requiring a large number of denoising steps to improve the accuracy of trajectory prediction. This often requires a significant amount of time, making it unfavorable for real-time trajectory prediction. We have learned the initial future trajectory distribution in the spatial-temporal interaction module, so only a few denoising steps are needed after sampling to achieve accurate prediction results. 

\begin{figure}[!h]
    \centering
    \subfigure[ADE]{\includegraphics[width=0.49\linewidth]{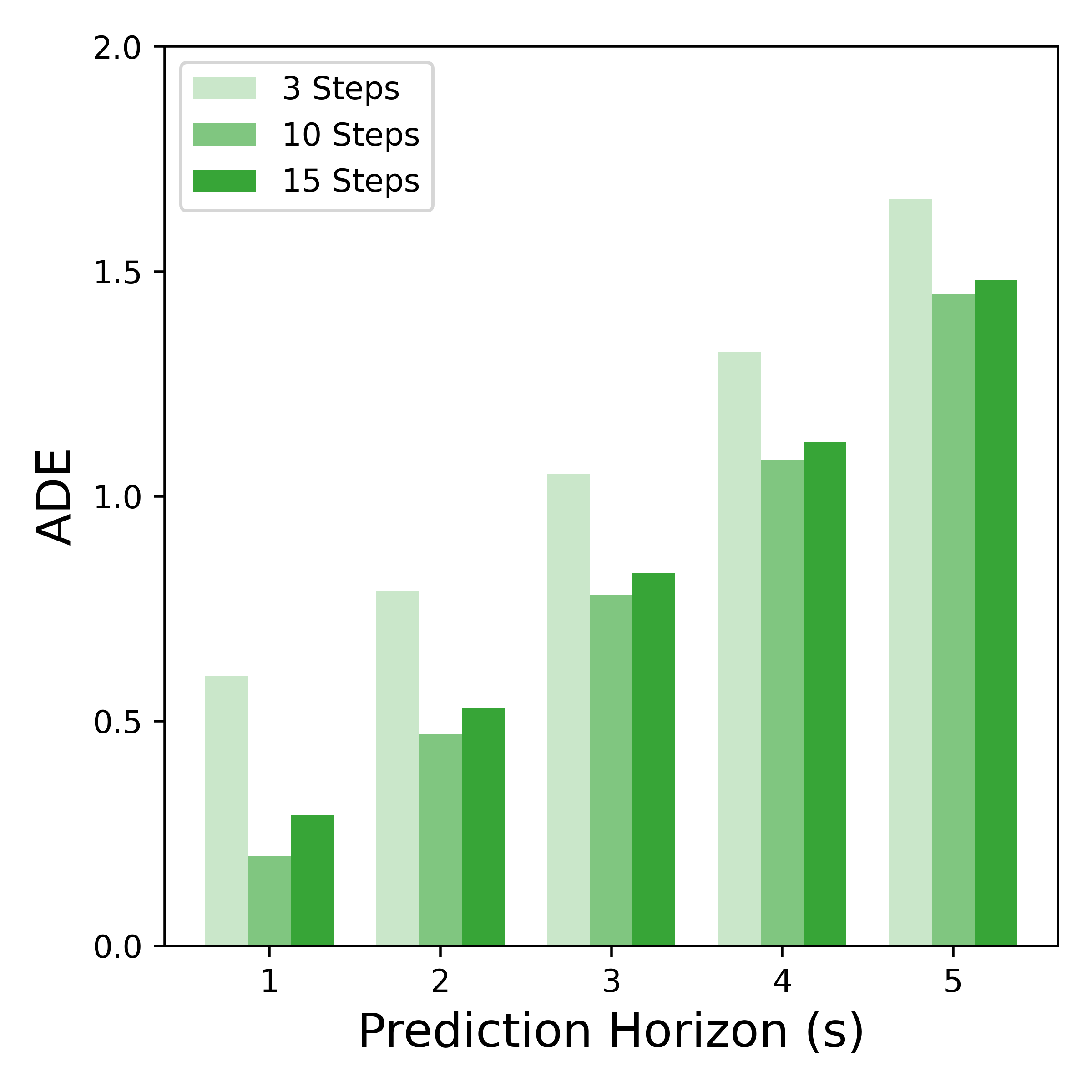}}
    \subfigure[FDE]{\includegraphics[width=0.49\linewidth]{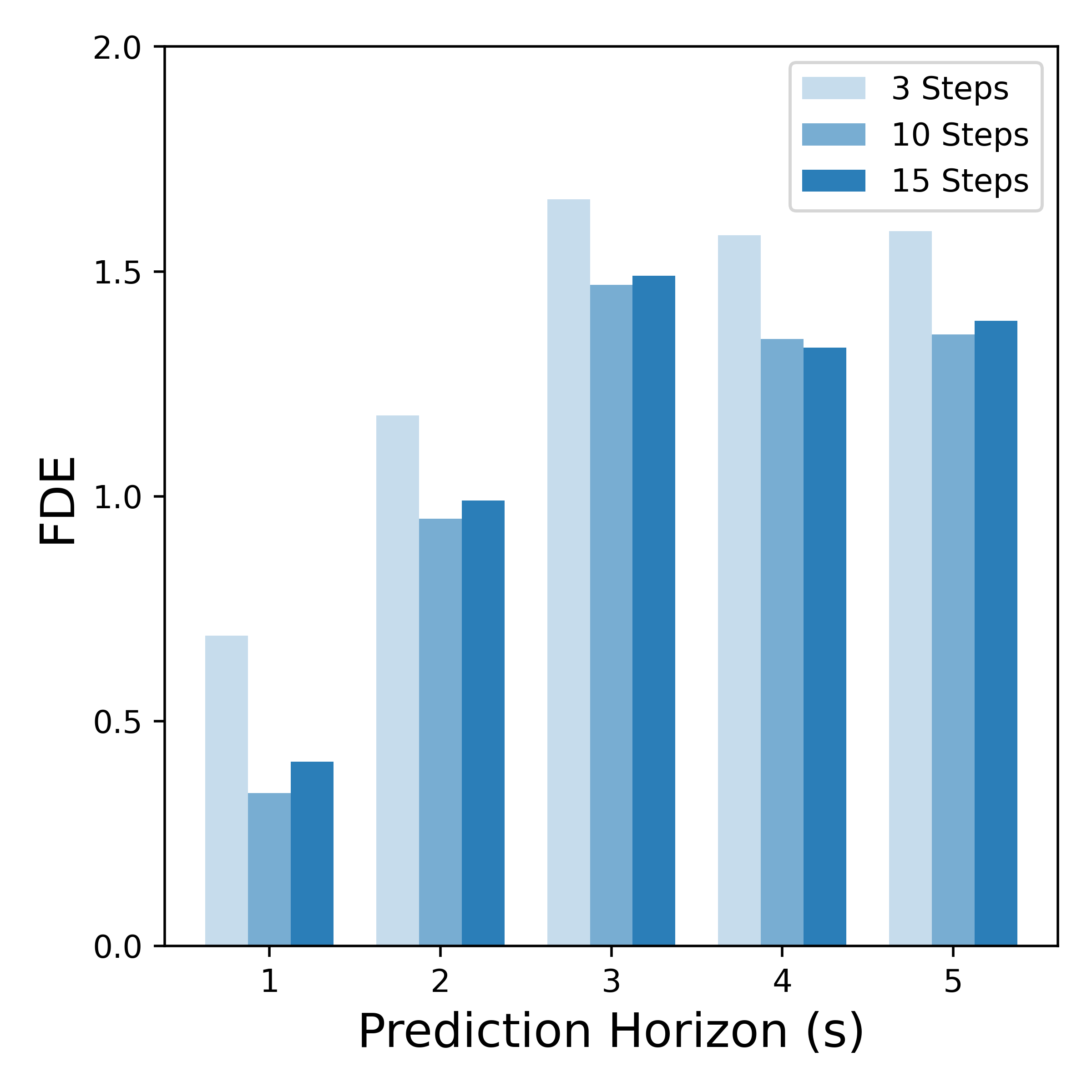}}
    \caption{The effect of denoising steps.}
    \label{fig:1}
\end{figure}

To achieve the best prediction results within the shortest prediction time, we conducted hyperparameter experiments on the number of denoising steps on the NGSIM dataset to find the optimal balance between prediction time and accuracy. The experimental results are shown in Table~\ref{time} and Figure~\ref{fig:1}, and it can be seen that we achieve the best balance of prediction time and accuracy at a denoising step of 10 steps.

\begin{table}[h]
\centering
\setlength{\tabcolsep}{7mm}
\begin{tabular}{ccc}
\toprule 
\specialrule{0em}{1pt}{1pt}
Dataset &Steps  & Inference (ms) \\
\specialrule{0em}{1pt}{1pt}
\midrule 
\specialrule{0em}{1pt}{1pt}
\multirow{3}{*}{NGSIM} &3 & 186\\
\specialrule{0em}{1pt}{1pt}
&10 &269\\
\specialrule{0em}{1pt}{1pt}
&15 &288\\
\specialrule{0em}{1pt}{1pt}
\bottomrule 
\end{tabular}
\caption{Inference time with different denoising steps.}
\label{time}
\end{table}

From Figure~\ref{fig:1}, it is evident that when the number of denoising steps is too small, the uncertainty in the trajectory cannot be fully eliminated, leading to a decline in performance. Conversely, when the number of denoising steps is too large, the refinement model has already achieved a sufficiently accurate trajectory, resulting in a performance bottleneck and unnecessary prediction time.

\subsection{Additional Case Study}
Figure~\ref{casea} shows an additional case study where three driving scenarios are selected including target keeping straight, merging to the left lane, and merging to the right lane. The observations are similar to that in our manuscript. Especially, C2F-TP can trace the right trajectory more accurately. Figure~\ref{caseb} shows that the target merges to the left lane. Both CS-LSTM and C2F-TP make correct direction predictions, but the predicted trajectory of C2F-TP is closer to the ground truth. Figure~\ref{casec} shows the target merges to the right lane. the predictions of S-LSTM and CS-LSTM keep going straight, while C2F-TP predicts that the target will merge into the right lane. This demonstrates that our method can more accurately predict future trajectories under different modalities.
\begin{figure}[!h]
    \centering
    \subfigure[Keeping straight]{\includegraphics[width=0.9\linewidth]{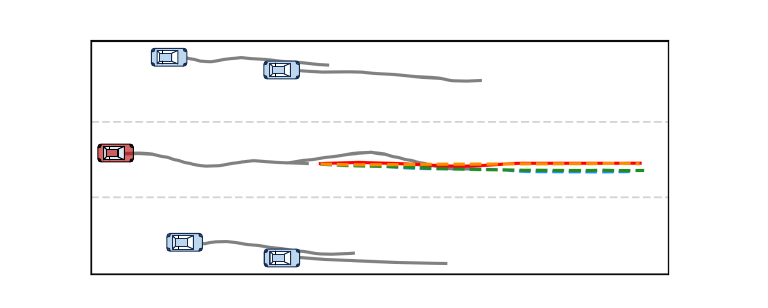}\label{casea}} 
    \subfigure[Merging to the left lane]{\includegraphics[width=0.9\linewidth]{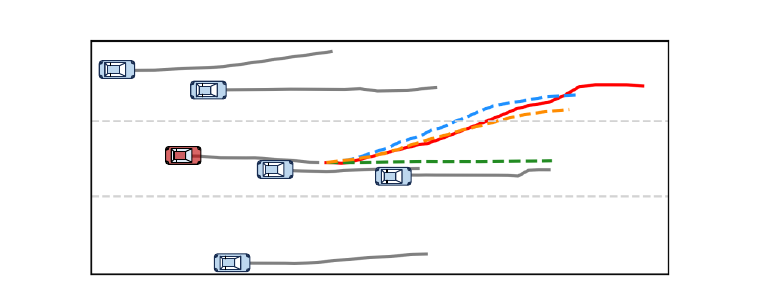}\label{caseb}}
    \subfigure[Merging to the right lane]{\includegraphics[width=0.9\linewidth]{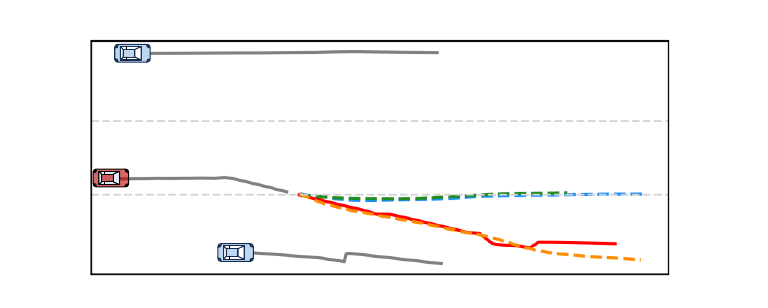}\label{casec}}
    \includegraphics[width=0.9\linewidth]{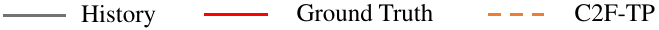}
    \includegraphics[width=1\linewidth]{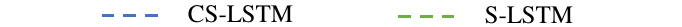}
    \caption{Visualisation of S-LSTM, CS-LSTM and C2F-TP for three driving scenarios.}
    \label{case}
\end{figure}

\end{document}